\def\year{2018}\relax
\documentclass[letterpaper]{article} %DO NOT CHANGE THIS
\usepackage{aaai18}  %Required
\usepackage{times}  %Required
\usepackage{helvet}  %Required
\usepackage{courier}  %Required
\usepackage{url}  %Required
\usepackage{graphicx}  %Required
\usepackage{multirow}
\usepackage[utf8x]{inputenc}
\begin{document}
\title{\emph{With 1 follower I must be AWESOME :P}.\\ Exploring the role of irony markers in irony recognition}

\author{
 Debanjan Ghosh$^{\mathsection}$ Smaranda Muresan$^{\ddagger}$\\
  $^{\mathsection}$School of Communication \&  Information, Rutgers University, NJ, USA \\
  $^{\ddagger}$Data Science Institute, Columbia University, NY, USA  \\
  {\fontsize{10}{1em}{\tt debanjan.ghosh@rutgers.edu, smara@columbia.edu}}}

\maketitle
\begin{abstract}
Conversations in social media often contain the use of irony or sarcasm, when the users say the opposite of what they really mean. Irony markers are the meta-communicative clues that inform the reader that an utterance is ironic. %Such markers expressing authors' sentiment, and opinions frequently occur in social media. In this paper,
 We propose a thorough analysis of theoretically grounded irony markers in two social media platforms: $Twitter$ and $Reddit$. Classification and frequency analysis show that for $Twitter$, typographic markers such as emoticons and emojis are the most discriminative markers to recognize ironic utterances, while for $Reddit$  the morphological markers (e.g., interjections, tag questions) are the most discriminative. %A genre-based study show users are more ironic in controversial topics (e.g., politics, religion) than sports or technology.  
\end{abstract}
%, particularly with the advent of word-embedding and artificial neural network techniques 

\section{Introduction}
With the advent of social media, irony and sarcasm detection has become an active area of research in Natural Language Processing (NLP) \cite{joshi2016automatic,riloff,joshi2015,ghosh2017role}.  Most computational studies have focused on building state-of-the-art models to detect whether an utterance or comment is ironic/sarcastic\footnote{We treat irony and sarcasm similarly in this paper.} or not, sometimes without theoretical grounding. 
%despite strong theoretical foundations in linguistics and psychology, a large number of empirical studies on irony detection are not theoretically motivated, with a few notable exceptions \cite{riloff,joshi2015,ghosh2017role}.
%SM add citation to Ghosh and Veale 
%Thus they treat irony detection as a standard binary classification task similar to any document/text classification problem. \footnote{We treat irony and sarcasm interchangeably.}
In linguistics and discourse studies, \citeauthor{attardo2000ironya} (2000) and later \citeauthor{burgers2010verbal} (2010) have studied two theoretical aspects of irony in the text: \emph{irony factors'} and  \emph{irony markers}. Irony factors are characteristics of ironic utterances that cannot be removed without destroying the irony. In contrast, irony markers are a meta-communicative clue that ``alert the reader to the fact that a sentence is ironical'' \cite{attardo2000ironya}. They can be removed and the utterance is still ironic.  

In this paper, we examine the role of irony markers in social media for irony recognition. Although punctuations, capitalization, and hyperboles are previously used as features in irony detection \cite{bamman2015contextualized,muresanjasist2016}, here we thoroughly analyze a set of theoretically-grounded types of irony markers, such as tropes (e.g., metaphors), morpho-syntactic indicators (e.g., tag questions), and typographic markers (e.g., emoji) and their use in ironic utterances. Consider the following two irony examples  from $Twitter$ and $Reddit$ given in Table \ref{table:examples}. 

\begin{table}
\centering
%\begin{tabular}{ |l|p{10cm}| } 
\begin{tabular}{ p{1.5cm}|p{6cm} }
\hline
%\multicolumn{1}{|c|}{Platform} & {c}{Turn Type} & 
%\multicolumn{1}{c|}{Turn pairs} \\
Platform  & \multicolumn{1}{c}{Utterances} \\
\hline
{$Reddit$} &  Are you telling me iPhone 5 is only marginally better than iPhone 4S? I thought we were reaching a golden age with this game-changing device. /s\\
{$Twitter$} & With 1 follower I must be AWESOME. :P  \#ironic \\
\hline 
\end{tabular}
\caption{ Use of irony markers in two social media platforms}
\label{table:examples}
\end{table}
Both utterances are labeled as ironic by their authors (using hashtags in $Twitter$ and the /s marker in $Reddit$). In the $Reddit$ example, the author uses several irony markers such as \emph{Rhetorical question} (e.g., ``are you telling'' \dots) and \emph{metaphor} (e.g., ``golden age''). In the $Twitter$ example, we notice the use of capitalization (``AWESOME'') and emoticons (``:P'' (tongue out)) that the author uses to alert the readers that it is an ironic tweet. 
%results?

We present three contributions in this paper. First, we provide a detailed investigation of a set of theoretically-grounded irony markers (e.g., tropes, morpho-syntactic, and typographic markers) in social media. We conduct the classification and frequency analysis based on their occurrence. Second, we analyze and compare the use of irony markers on two social media platforms  ($Reddit$ and $Twitter$). Third, we provide an analysis of markers on topically different social media content (e.g., technology vs. political subreddits). 

\section{Data}
%Social media data from micro-blogging platform: $Twitter$ and discussion forum: $Reddit$ are used here.
\emph{\textbf{Twitter:}} We use a set of 350K tweets for our experiments. The ironic/sarcastic tweets are collected
using hashtags, such as \#irony, \#sarcasm, and \#sarcastic whereas the non-sarcastic tweets do not contain these hashtags, but they
might include sentiment hashtags, such as \#happy,
\#love, \#sad, \#hate (similar to \cite{gonzalez,ghoshguomuresan2015EMNLP}). As pre-processing, we removed the retweets, spam, duplicates, and tweets written in languages other than English. Also, we deleted all tweets where the hashtags of interest were not located at the very end (i.e., we eliminated ``\#sarcasm is something that I love''). We lowercased the tweets, except the words where all the characters are uppercased. 

%different example?
\emph{\textbf{Reddit:}} \citeauthor{khodak2017large} (2018) introduced an extensive collection of sarcastic and non-sarcastic posts collected from different subreddits. In Reddit, authors mark their sarcastic  intent of their posts by  adding ``/s''  at the end of a post/comment. We collected 50K instances from the corpus for our experiments (denoted as $Reddit$), where the sarcastic and non-sarcastic replies are at least two sentences (i.e., we discard posts that are too short).% \citeauthor{manningtool} (2014) is used to $Reddit$ the forum posts.    

For brevity, we denote ironic utterances as $I$ and non-ironic utterances as $NI$. Both $Twitter$ and $Reddit$ datasets are balanced between the $I$ and $NI$ classes. We uuse 80\% of the datasets for training, 10\% for development, and the remaining 10\% for testing.

\section{Irony Markers}
Three types of markers --- tropes, morpho-syntactic, and typographic are used as features. %Note, although some markers, such as hyperbole and capitalization features are used in related literature \cite{bamman2015contextualized}, we provide a detailed systematic analysis of a larger set of markers here. 

\subsection{Tropes:}
Tropes are figurative use of expressions.
\begin{itemize}
\item Metaphors - Metaphors often facilitate ironic representation and are used as markers. We have drawn metaphors from different sources (e.g., 884 and 8,600 adjective/noun metaphors from \cite{tsvetkov2014metaphor} and \cite{gutierrez2016literal}, respectively, and used them as binary features. We also evaluate the metaphor detector \cite{rei2017grasping} over $Twitter$ and $Reddit$ datasets. We considered metaphor candidates that have precision $\ge$ 0.75 (see \citeauthor{rei2017grasping} (2017)). 
\item Hyperbole - Hyperboles or intensifiers are commonly used in irony because speakers frequently overstate the magnitude of a situation or event. We use terms that are denoted as ``strong subjective'' (positive/negative) from the MPQA corpus \cite{wilson2005recognizing} as hyperboles. Apart from using hyperboles directly as the  binary feature we also use their sentiment as features. 
\item Rhetorical Questions - Rhetorical Questions (for brevity $RQ$) have the structure of a question but are not typical information seeking questions. We follow the hypothesis introduced by \citeauthor{oraby2017you} (2017) that questions that are in the middle of a comment are more likely to be RQ since since questions followed by text cannot be typical information seeking questions. Presence of $RQ$ is used as a binary feature. 
\end{itemize}
\subsection{Morpho-syntactic (MS) irony markers:}
This type of markers appear at the morphologic and syntactic levels of an utterance.
\begin{itemize}
\item Exclamation - Exclamation marks emphasize a sense of surprise on the literal evaluation that is reversed in the ironic reading \cite{burgers2010verbal}. We use two binary features, single or multiple uses of the marker.

\item Tag questions -  We built a list of tag questions (e.g.,, ``didn't you?'', ``aren't we?'') from a grammar site and use them as binary indicators.\footnote{http://www.perfect-english-grammar.com/tag-questions.html}

\item Interjections - Interjections seem to undermine a literal evaluation and occur frequently in ironic utterances (e.g., ```yeah", `wow'', ``yay'',``ouch'' etc.). Similar to tag questions we assembled interjections (a total of 250) from different grammar sites.  
\end{itemize}

\subsection{Typographic irony markers:}

\begin{itemize}
\item Capitalization - Users often capitalize words to represent their ironic use (e.g., the use of ``GREAT", ``SO'', and ``WONDERFUL'' in the ironic tweet ``GREAT i'm SO happy shattered phone on this WONDERFUL day!!!''). 

\item Quotation mark - Users regularly put quotation marks to stress the ironic meaning (e.g., ``great'' instead of GREAT in the above example).

\item Other punctuation marks -  Punctuation marks such as ``?'', ``.'', ``;'' and their various uses (e.g., single/multiple/mix of two different punctuations) are used as features.

\item Hashtag - Particularly in $Twitter$, hashtags often represent the sentiment of the author. For example, in the ironic tweet ``nice to wake up to cute text. \#suck'', the hashtag ``\#suck'' depicts the negative sentiment. We use binary sentiment feature  (positive or negative) to identify the sentiment of the hashtag, while comparing against the MPQA sentiment lexicon. Often multiple words are combined in a hashtag without spacing (e.g., ``fun'' and ``night'' in \#funnight). We use an off-the-shelf tool to split words in such hashtags and then checked the sentiment of the words.\footnote{https://github.com/matchado/HashTagSplitter}

\item Emoticon - Emoticons are frequently used to emphasize the ironic intent of the user. In the example ``I love the weather ;) \#irony'', the emoticon ``;)'' (wink) alerts the reader to a possible ironic interpretation of weather (i.e., bad weather). We collected  a comprehensive list of emoticons (over one-hundred) from Wikipedia and also used standard regular expressions to identify emoticons in our datasets.\footnote{http://sentiment.christopherpotts.net/code-data/} Beside using the emoticons directly as binary features, we use their sentiment as features as well (e.g., ``wink'' is regarded as positive sentiment in MPQA). 

\item Emoji - Emojis are like emoticons, but they are actual pictures and recently have become very popular in social media. Figure \ref{figure:emoji} shows a tweet with two emojis (e.g., ``unassumed'' and ``confounded'' faces respectively) used as markers. We use an emoji library of 1,400 emojis to identify the particular emoji used in irony utterances and use them as binary indicators.\footnote{https://github.com/vdurmont/emoji-java}
\begin{figure}[t]
\center
\includegraphics[width=3in]{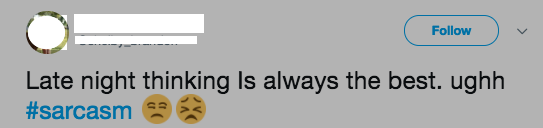}

\caption{Utterance with emoji (best in color)}
\label{figure:emoji}
\end{figure}

\end{itemize}

\section{Classification Experiments and Results}

We first conduct a binary classification task to decide whether an utterance (e.g., a tweet or a $Reddit$ post) is ironic or non-ironic, exclusively based on the irony marker features. We use Support Vector Machines (SVM) classifier with linear kernel  \cite{fan2008liblinear}. %with category weights inversely proportional to utterance frequencies. 
%SM I remoed the weights as is balanced no? 
Table \ref{table:tweetresults} and Table \ref{table:redditresults} present the results of the ablation tests for $Twitter$ and $Reddit$. We report Precision ($P$), Recall ($R$) and $F1$ scores of both $I$ and $NI$ categories.  

\begin{table}
\centering
\begin{small}
\begin{tabular}{lcccc}
\hline
Features & Category  & $P$ & $R$ & $F1$ \\
\hline
%\multirow {2} {*} {all} & $I$  & 66.93 & \textbf{77.32} & \textbf{71.75} \\
%& $NI$ & \textbf{73.13} & 61.78 & \textbf{66.97} \\
%\hline
\multirow {2} {*} {all} & $I$  & 66.93 & \textbf{77.32} & \textbf{71.75} \\
& $NI$ & \textbf{73.13} & 61.78 & \textbf{66.97} \\
\hline
\multirow {2} {*} {- tropes} & $I$  & \textbf{67.70} & {48.00} & 56.18 \\
& $NI$ & {59.70} & \textbf{77.09} & \textbf{67.29} \\
\hline
\multirow {2} {*} {- MS} & $I$  & 63.59 & \textbf{78.09} & 70.10\\
& $NI$ & \textbf{71.59} & {55.27} & 62.38 \\
\hline
\multirow {2} {*} {- typography} & $I$  & 57.30 & 77.95 & 66.05 \\
& $NI$ & 65.49 & 41.86 & 51.07 \\
\hline
\end{tabular}
\end{small}
\caption{Ablation Tests of irony markers for $Twitter$. \textbf{bold} are best scores (in \%).}
%\centering
\label{table:tweetresults}
\end{table}

\begin{table}
\centering
\begin{small}
\begin{tabular}{lcccc}
\hline
Features & Category  & $P$ & $R$ & $F1$ \\
\hline
\multirow {2} {*} {all} & $I$  & \textbf{73.16} & {48.52} & \textbf{58.35} \\
& $NI$ & \textbf{61.49} & \textbf{82.20} & \textbf{70.35} \\
\hline
\multirow {2} {*} {- tropes} & $I$  & {71.45} & \textbf{50.36} & \textbf{59.08} \\
& $NI$ & \textbf{61.67} & {79.88} & \textbf{69.61} \\
\hline
\multirow {2} {*} {- MS} & $I$  & 58.37 & \textbf{49.36} & 53.49\\
& $NI$ & {56.13} & {64.8} & 60.16 \\
\hline
\multirow {2} {*} {- typography} & $I$  & 73.29 & 48.52 & 58.39 \\
& $NI$ & \textbf{61.52} & \textbf{82.32} &  \textbf{70.42} \\
\hline
\end{tabular}
\end{small}
\caption{Ablation Tests of irony markers for $Reddit$ posts. \textbf{bold} are best scores (in \%).}
%\centering
\label{table:redditresults}
\end{table}

Table \ref{table:tweetresults} shows that for ironic utterances in $Twitter$, removing tropes have the maximum negative effect on Recall, with a reduction on $F1$ score by 15\%. This is primarily due to the removal of hyperboles that frequently appear in ironic utterances in $Twitter$. Removing typographic markers (e.g., emojis, emoticons, etc.) have the maximum negative effect on the Precision for the irony $I$ category, since particular emojis and emoticons appear regularly in ironic utterances (Table \ref{table:tweettopf}). For $Reddit$, Table \ref{table:redditresults} shows that removal of typographic markers such as emoticons does not affect the F1 scores, whereas the removal of morpho-syntactic markers, e.g., tag questions, interjections have a negative effect on the F1. %Note, a strong lexical baseline (e.g., unigram and bigram features) results in 75\% and 60\% F1 for $Twitter$ and $Reddit$, respectively, which show irony markers independently are comparable features for irony recognition task.  
%On an average, removal of typographic markers influence the drop of F1 scores for both the categories (i.e., for $NI$, F1 reduces by 15\%).

Table \ref{table:tweettopf} and Table \ref{table:reddittopf} represent the $top$ most discriminative features for both categories based on the feature weights learned during the SVM training for $Twitter$ and $Reddit$, respectively. Table \ref{table:tweettopf} shows that for $Twitter$, typographic features such as emojis and emoticons have the highest feature weights for both categories. Interestingly, we observe that for ironic tweets users often express negative sentiment directly via emojis (e.g., angry face, rage) whereas for non-ironic utterances, emojis with positive sentiments (e.g., hearts, wedding) are more familiar. For $Reddit$ (Table \ref{table:reddittopf}), we observe that instead of emojis, other markers such as exclamation marks, negative tag questions, and metaphors are discriminatory markers for the irony category. In contrary, for the non-irony category, positive tag questions and negative sentiment hyperboles are influential features.      
\begin{table}
\centering
\begin{small}
\begin{tabular}{p{1cm}p{7cm}}
\hline
Category & Top features  \\
\hline
$I$  & emoticons: annoyed (``-\_-''), perplexed (``:-/''); emojis: angry face/monster, unamused, expressionless, confounded, rage, neutral\_face, thumbsdown; negative\_tag questions (``is n't it?'', ``don't they?")   \\
\hline
$NI$ & emojis: birthday, tophat, hearts, wedding, rose, ballot\_box\_with\_check; quotations, hashtags (positive sentiment), emoticons: happy (``:)''), overjoyed (``$\wedge\_\wedge$'')  \\
\hline
\end{tabular}
\end{small}
\caption{Irony markers based on feature weights for $Twitter$}
%\centering
\label{table:tweettopf}
\end{table}
\begin{table}
\centering
\begin{small}
\begin{tabular}{p{1cm}p{7cm}}
\hline
Category & Top features  \\
\hline
$I$  & exclamation (single, multiple), negative\_tag questions (``is n't it?'', ``don't they?"), interjections, presence of metaphors, positive sentiment hyperbolic words (e.g., ``notably'', ``goodwill'', ``recommendation'')   \\
\hline
$NI$ & negative sentiment hyperbolic words (e.g., ``vile'', ``lowly'', ``fanatic''), emoticon: laugh (``:))''), positive\_taq questions (``is it?'', ``are they?''), punctuations such as periods/multiple periods  \\
\hline
\end{tabular}
\end{small}
\caption{Irony markers based on feature weights for $Reddit$}
%\centering
\label{table:reddittopf}
\end{table}

\begin{table*}
\centering
\begin{small}
%\begin{tabular}{cccc}
\begin{tabular}{p{1.5cm}p{2cm}p{2cm}p{2cm}p{2.6cm}p{2.6cm}}
\hline
\multicolumn{2}{c}{Irony Markers} & \multicolumn{4}{c}{Genres} \\

Type & Marker & Technology (a) & Sports (b) & Politics (c) & Religion (d) \\
\hline
& Metaphor & 0.01 (0.06) & 0.002 (0.05) & 0.02 (0.12) & 0.01 (0.10)\\
Trope & Hyperbole & 0.19 (0.39) &  0.34 (0.48)$^{a{^{**}}}$ & 0.74 (0.44)$^{(a,b){^{**}}}$  & 0.76 (0.43)$^{{(a,b)^{**}}, c{^*}}$\\
& $RQ$ &  0.06 (0.23) & 0.11 (0.32)$^{a{^{**}}}$ & 0.22 (0.41)$^{(a,b){^{**}}}$ & 0.2 (0.4)$^{(a,b){^{**}}}$\\
\hline
& Exclamation & 0.09 (0.29) & 0.14 (0.34)$^{a{^{**}}}$ & 0.42 (0.49)$^{(a,b){^{**}}}$ & 0.37 (0.48)$^{(a,b,c){^{**}}}$\\
MS & Tag Question &  0.03 (0.16) & 0.05 (0.23)$^{a{^{**}}}$ & 0.11 (0.32)$^{(a,b){^{**}}}$ & 0.1 (0.30)$^{(a,b){^{**}}}$ \\
& Interjection &  0.13 (0.34) & 0.23 (0.42)$^{a{^{**}}}$ & 0.45 (0.50)$^{(a,b){^{**}}}$ & 0.52 (0.5)$^{(a,b,c){^{**}}}$\\
\hline
& Capitalization &   0.04 (0.19) & 0.08 (0.27)$^{a{^{**}}}$ & 0.20 (0.40)$^{(a,b){^{**}}}$ & 0.1 (0.31)$^{(a,b,c){^{**}}}$\\
%& Quotation &  - & - \\
Typographic & Punctuations &  0.23 (0.42) & 0.45 (0.50)$^{a{^{**}}}$ & 0.84 (0.36)$^{(a,b){^{**}}}$ & 0.89 (0.31)$^{(a,b,c){^{**}}}$\\
%& Hashtag &  - &  - \\
%& Emoticon &  - & - & 0.01 (0.04) & - \\
%& Emoji &  - & - \\
\hline
\end{tabular}
\end{small}
\caption{Frequency of irony markers in different genres (subreddits). The mean and the SD (in bracket) are reported.$^{x{^{**}}}$ and $^{x{^{*}}}$ depict significance on $p\le 0.005$ and $p\le 0.05$, respectively.}
%\centering
\label{table:meangenre}
\end{table*}
\subsection{Frequency analysis of markers}
We also investigate the occurrence of markers in the two platforms via frequency analysis (Table \ref{table:freq}). We report the mean of occurrence per utterance and the standard deviation (SD) of each marker. Table \ref{table:freq} shows that markers such as hyperbole, punctuations, and interjections are popular in both platforms. Emojis and emoticons, although the two most popular markers in $Twitter$ are almost unused in $Reddit$. Exclamations and $RQ$s are more common in the $Reddit$ corpus. Next, we combine each marker with the type they belong to (i.e., either trope, morpho-syntactic and typographic) and compare the means between each pair of types via independent t-tests. We found that the difference of means is significant ($p \leq 0.005$) for all pair of types across the two platforms. 

\begin{table}
\centering
\begin{small}
%\begin{tabular}{cccc}
\begin{tabular}{p{1.5cm}p{1.8cm}p{1.8cm}p{1.8cm}}
\hline
\multicolumn{2}{c}{Irony Markers} & \multicolumn{2}{c}{Corpus} \\

Type & Marker & $Twitter$ & $Reddit$ \\
\hline
& Metaphor & 0.02 (0.16) & 0.01 (0.08)\\
Trope & Hyperbole & 0.45 (0.50) &  0.43 (0.50)\\
& $RQ$ &  0.01 (0.08) & 0.15 (0.36)\\
\hline
& Exclamation & 0.02 (0.16) & 0.19 (0.39)\\
MS & Tag Question &  0.02 (0.10) & 0.08 (0.26) \\
& Interjection &  0.22 (0.42) & 0.32 (0.46) \\
\hline
& Capitalization &   0.03 (0.16) & 0.10 (0.30)\\
& Quotation &  0.01 (0.01) & - \\
Typographic & Punctuations &  0.10 (0.29) & 0.47 (0.50)\\
& Hashtag &  0.02 (0.14) &  - \\
& Emoticon &  0.03 (0.14) & 0.001  (0.03) \\
& Emoji &  0.05 (0.22) & - \\
\hline
\end{tabular}
\end{small}
\caption{Frequency of irony markers in two platforms. The mean and the SD (in bracket) are reported.}
%\centering
\label{table:freq}
\end{table}

\subsection{Irony markers across topical subreddits}
Finally, we collected another set of irony posts from \cite{khodak2017large}, but this time we collected posts from specific topical subreddits. We collected irony posts about politics (e.g., subreddits: politics, hillary, the\_donald), sports (e.g., subreddits: nba, football, soccer), religion (e.g., subreddits: religion) and technology (e.g., subreddits: technology). Table \ref{table:meangenre} presents the mean and SD for each genre. We observe that users use tropes such as hyperbole and $RQ$, morpho-syntactic markers such as exclamation and interjections and multiple-punctuations more in politics and religion than in technology and sports. This is expected since subreddits regarding politics and religion are often more controversial than technology and sports and the users might want to stress that they are ironic or sarcastic using the markers. %Thus we reckon users on these subreddits are more opinionated and frequent in being sarcastic and ironic towards others.

\section{Conclusion}
We provided a thorough investigation of irony markers across two social media platforms: Twitter and Reddit. Classification experiments and frequency analysis suggest that typographic markers such as emojis and emoticons are most frequent for $Twitter$ whereas tag questions, exclamation, metaphors are frequent for $Reddit$. We also provide an analysis across  different topical subreddits. In future, we are planning to experiment with other markers (e.g., ironic echo, repetition, understatements).

\iffalse
!_MUL: 0.707460
!: 0.551737
?_MUL: 0.375296
?: 0.350587
NEGATIVE_TAG_PRESENT: 0.118520
EMOTI_smiley_PRESENT: 0.046256
INTERJ_PRESENT: 0.011863
EMOTI_POSITIVE_PRESENT: 0.009792
METAPHOR_PRESENT: 0.005127
HYPERBOLIC_PRESENT: 0.004167

HYPERBOLIC_NEGATIVE_PRESENT: -0.004330
TAG_PRESENT: -0.015489
UPPER_PRESENT: -0.022623
EMOTI_laugh_PRESENT: -0.036464
.: -0.078927
HYPERBOLIC_POSITIVE_PRESENT: -0.105323
POSITIVE_TAG_PRESENT: -0.134009
RQ_PRESENT: -0.140100
._MUL: -0.319297
\fi

%\bibliography{markers_bib}

\bibliographystyle{aaai}
\end{document}